\documentclass{article}
\usepackage{spconf,amsmath,graphicx}
\usepackage{subfigure}
\usepackage{hyperref}
\usepackage{multicol, multirow}
\usepackage{amsthm,amsmath,amssymb}
\usepackage{cleveref}
\usepackage{stfloats}
\usepackage{tikz}
\usepackage{pgfplots}
\usepackage{graphics}
\crefname{section}{§}{§§}
\Crefname{section}{§}{§§}

\usepackage{mathrsfs}
\usepackage{pgfplots}
\pgfplotsset{compat=1.17}

\title{PAGE: A POSITION-AWARE GRAPH-BASED MODEL FOR EMOTION CAUSE ENTAILMENT IN CONVERSATION}

\name{Xiaojie Gu$^{1}$, Renze Lou$^{2}$, Lin Sun$^{1}$\sthanks{Corresponding author.}, Shangxin Li$^{1}$}
\address{$^{1}$Department of Computer Science, Hangzhou City University, Hangzhou, China\\
$^{2}$ Department of Computer Science and Engineering, Pennsylvania State University, State College, USA}

\begin{document}
%
\maketitle

\begin{abstract}
\textbf{C}onversational \textbf{C}ausal \textbf{E}motion \textbf{E}ntailment (${\rm C}_2{\rm E}_2$)
is a task that aims at recognizing the causes corresponding to a target emotion in a conversation.
The order of utterances in the conversation affects the causal inference.
However, most current position encoding strategies ignore the order relation among utterances and speakers.
To address the issue, we devise a novel position-aware graph to encode the entire conversation, fully modeling causal relations among utterances.
The comprehensive experiments show that our method consistently achieves state-of-the-art performance on two challenging test sets, proving the effectiveness of our model.
Our source code is available on Github\footnote{\url{https://github.com/XiaojieGu/PAGE}}.
\end{abstract}
\section{Introduction}
\label{sec:intro}

For a target utterance transmitting a specific emotion,
the ${\rm C}_2{\rm E}_2$ aims to identify the causal utterances from the conversation history responsible for the target emotion.
This novel task is essential to design current dialogue agents, such as empathetic response \cite{shin2020generating}
and emotion counseling \cite{liu2021towards}.
Besides, it provides a potential way to improve the interpretability of affect-based models \cite{poria2021recognizing}. 

Lee et al.~\cite{lee2010text} first proposed emotion cause extraction (ECE), who pointed out the importance of this task and considered it in a sentence-level classification paradigm.
Early research employed rule-based strategies \cite{chen2010emotion} and traditional machine learning approaches \cite{ghazi2015detecting,russo2011emocause} to deal with this problem.
As a further step toward the ECE, ${\rm C}_2{\rm E}_2$ considers a more challenging conversation scenario.
So far, only a few studies \cite{poria2021recognizing,li2022neutral} have put the finger on ${\rm C}_2{\rm E}_2$.
Porial et al. \cite{poria2021recognizing} solely paired one target utterance with other utterances,
which loses contextual information and breaks the causal relationship between utterances.
Li et al. \cite{li2022neutral} utilized commonsense knowledge to facilitate causes recognization.
From our insight view,
the contextual information has a significant effect on utterance understanding,
and it is difficult to determine whether they have a corresponding causality without considering the position between them \cite{ding2020ecpe,yan2021position}.
For that, a naive way is to directly concatenate the absolute position embedding with the utterance representation \cite{ding2019independent,li2019context}.
However, this scheme constantly leads to the aggregation of uninformative context, bringing inference noise (i.e., the causal-irrelevant context).
To this end, Xia et al. \cite{xia2019rthn} utilized a multi-head attention mechanism to weigh the position information.
Ding et al. \cite{ding2020ecpe} set the window size to consider only the adjacency of target utterances.
While these position encoding strategies target the documents rather than conversations.
Therefore, they neglected the vital inter-speaker dependency, which is essential in understanding conversation \cite{ghosal2019dialoguegcn}. 

To address the above issues,
we propose PAGE (\textbf{P}osition-\textbf{A}ware \textbf{G}raph-based model for \textbf{E}motion cause entailment),
in which we devise a novel relative position encoding schema to distinguish utterances of different speakers for better reasoning.
Intuitively, relative position plays a vital role in conversation-based causal inference.
For example, there are two causal-relevant utterances from distinct speakers, namely ``\emph{Hey, you wanna see a movie tomorrow?}'' and ``\emph{Sounds like a good plan.}''.
If the order is reversed, we probably do not realize that ``a good plan'' refers to ``see a movie''.
Therefore, we construct the position relationship based on the relative distance between the different utterances of the inter-speaker.
Furthermore, there is an explicit topological relationship between the emotion and the cause \cite{Moors2010-MOOACA-3},
so we leverage graph neural networks to encode the entire conversation context.
We evaluate our approach on the latest benchmark dataset. The experimental results show that our
method gains competitive performance compared with other strong baselines.
In summary, the main contributions of this paper are as follows:

$\bullet$ We propose a position encoding strategy that can enhance emotion cause entailment and the understanding of conversation context in ${\rm C}_2{\rm E}_2$.
We design a novel position-aware graph to better aggregate the entire conversation.

$\bullet$ 
We conduct comprehensive experiments to demonstrate the effectiveness of our approach and provide a thorough analysis.
Our model achieves state-of-the-art performance on the benchmark dataset, especially improving 8.1\% absolute in F1 score on the IE test set.

\begin{figure*}[ht]
	\begin{center}
		\centering
		\includegraphics[width=1.0\linewidth]{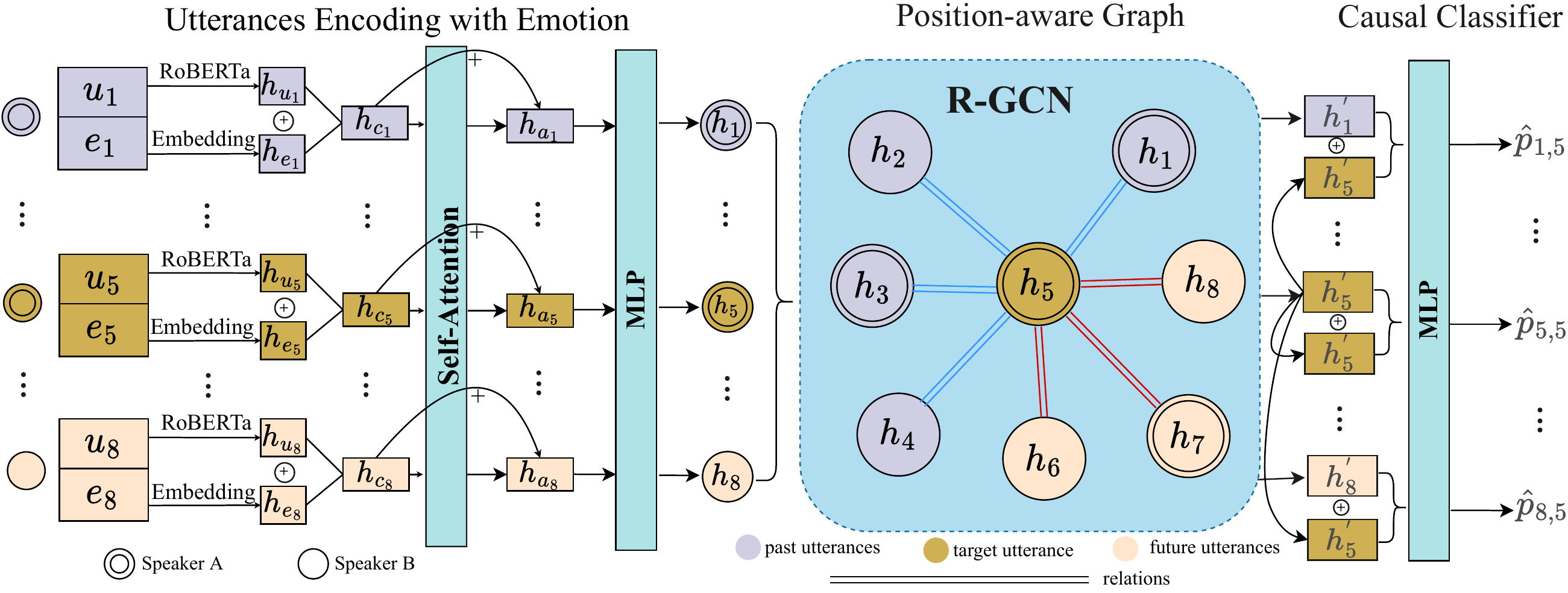}
    	\end{center}
	\caption{Overall architecture of the PAGE.}
	\label{fig:architectures}
\end{figure*}

\section{METHODOLOGY}
As shown in Figure \ref{fig:architectures}, our framework consists of three components, namely Utterances Encoding with Emotion (\cref{conv_enc}), Position-aware Graph (PaG) (\cref{posi_enc}) and Causal Classifier (\cref{cause_enc}).
Given a conversation $C = \{u_1,u_2,\ldots,u_k\}$ where there are target utterances $u_t \in C $
of speaker $S_t \in \{A,B\}$.
Each $u_t$ is labeled with a specific non-neutral emotion (e.g., \textit{``happiness'', ``anger'', `` sadness''}).
Our goal is to identify the causal utterances in the conversation history of \(u_t\).
That is, whether a \(u_i\) ( \(i \leq t\)) is a cause for \(u_t\) or not.

\subsection{Utterances Encoding With Emotion}\label{conv_enc}
For an utterance $u_n = \{w_1,w_2,\ldots,w_m\}$ consisting of $m$ words,
we add two special tokens $[CLS]$ and $[SEP]$, at the beginning and end of it, respectively.
Then, we use Roberta \cite{liu2019roberta} to conduct sentence-level encoding
and take the hidden state of the last layer as word-level representation:
\begin{align}
h_w = \textrm{RoBERTa}\left([CLS],w_1,w_2,\ldots,w_{m},[SEP]\right),
\end{align}
where $h_w \in \mathbb{R}^{m \times {d_{R}}}$ and $d_{R} = 768$ denotes the dimension of word-level representation in RoBERTa.
Then, we get utterance representation $h_u \in \mathbb{R}^{d_u} $ by a linear projection $W_u \in \mathbb{R}^{d_{R} \times d_u}$ on $h_w$.

Emotion information in the conversation can pass among speakers \cite{song2022emotionflow},
which is beneficial for entailment \cite{hu2021bidirectional};
thus,
we concatenate emotion embedding with the utterance representation, i.e., $h_c = h_e \oplus h_u$.
To capture the utterance features from multiple aspects,
we employ a multi-head self-attention mechanism \cite{vaswani2017attention},
which contributes to sentence-level sentiment analysis \cite{lin2020self}.
The value of the Q, K, and V vectors are the same as $h_c$:
\begin{align}
head_N = \textrm{softmax}\left(\frac{QK^T}{\sqrt{d_u}}\right)V,
\end{align}
where $head_N \in \mathbb{R}^{\frac{d_u}{N}}$ and N is the number of head.
We concatenate heads together to get the attention output \(h_a\) and add it to $h_c$ by $x = h_a +h_c$.
Then, followed by multilayer perceptron (MLP) with a single hidden layer and a residual connection:
\begin{align}
h_n&= \sigma\left(MLP(x)\right)+x,
\end{align}
where the output dimension of the MLP is $\mathbb{R}^{d_u}$ and $ \sigma(\cdot)$ is a sigmoid function.

\subsection{Position-aware Graph}\label{posi_enc}
To alleviate long-term dependency between utterances in long conversations,
we utilize graph neural networks to perform position-aware encoding.
We design a directed graph denoted as $\mathcal{G} = (\mathcal{V},\mathcal{E},\mathcal{R})$.
The utterance $u_n \in \mathcal{V}$ in the conversation denotes a node,
whose initial representation is $h_n$,
and $r_{o,t} \in \mathcal{R}$ is the type of an edge $(u_o,r_{o,t},u_t) \in \mathcal{E}$,
where $u_t$ denotes the target utterance and $u_o$ ( \(1 \leq o \leq n\) ) denotes other utterances.

The relative position plays a significant role in the transformation of causal information between utterances; 
thus, we represent the type of edges between nodes in relative position relations.
%
Moreover, the sequence relationship between the utterances of the same or different speakers facilitates the information understanding in the utterances and enhances entailment.
For this, we construct a positional relation $r_{o,t}$ based on a relative distance $D_{o,t}$ that denotes the distance between the target utterance $u_t$ and a neighboring utterance $u_o$:
\begin{eqnarray}
D_{o,t}=
\begin{cases}
\frac{o-t}{2} & {S_o=S_t}\\
-1 & {S_o \neq S_t\ and\ t=o \pm 1}.\\
\frac{o-t-1}{2} & {others}
\end{cases} 
\end{eqnarray}

\begin{figure}[t]
	\begin{center}
		\centering
		\scalebox{0.7}{
		\includegraphics[width=1.0\linewidth]{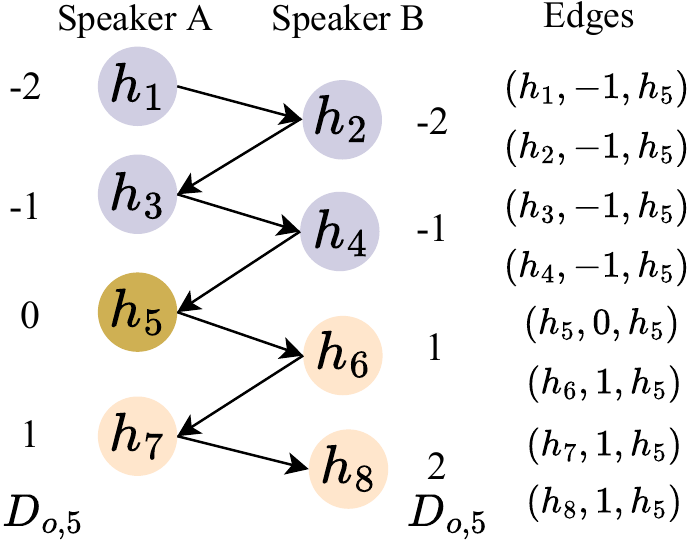} }
	\end{center}
	\caption{An example of our relative position encoding. The target utterance is $u_5$ and window size $w=1$. Purple represents past utterances, orange represents target utterance and cream represents future utterances.}
	\label{fig:relation}
\end{figure}

As the example shown in Figure \ref{fig:relation}, the target utterance $u_5$ has the same relative distance from both surrounding utterances $u_3$ and $u_4$ in the case of the same or different speakers, i.e., $D_{3,5} = D_{4,5}$,
they are considered to be in the same positional relationship, i.e., $r_{3,5} = r_{4,5}$.
For the utterance $u_j$ where $j > t$, the utterances after the target utterance are not dominant compared to the parts before the target utterance \cite{li2021past},
but they still can serve as additional context information to conversation understanding.
We cannot completely ignore them, so we treat the parts after the target utterance as the same relative position relation.
In this paper, we set it to 1.
Furthermore, previous work indicates that most causes are located near the emotion \cite{yan2021position},
and if two utterances are far apart,
then the causal relationship between them would diminish.
We do not need to give them the same attention compared to those near the target utterance.
To alleviate this issue, we set a window size in which
the relative position relation between $u_t$ and $u_j$ is set to the same.
In short, we get relative position relation $r_{o,t}$ from $u_o$ to $u_t$ by:
\begin{eqnarray}
r_{o,t}=
\begin{cases}
-w & {D_{o,t} < -w}\\
D_{o,t} & {D_{o,t} \geq -w\ and\ o \leq t}    .\\
1 & {o > t}
\end{cases}
\end{eqnarray}

We use R-GCN \cite{schlichtkrull2018modeling}, which can integrate different relationships between nodes 
to get the final utterance representation which is beneficial to the position-aware transformation:
\begin{align}
h_{t}^{'} = \sigma\left(\sum_{r \in \mathcal{R}}\sum_{o \in \mathcal{N}_{t}^{r}}\frac{1}{c_{t,r}}W_{r}h_{o}+W_{0}h_{t}
\right),
\end{align}
where $\mathcal{N}_{t}^{r}$ denotes the set of neighboring nodes of node $t$ under the relationship $r$,
$c_{t,r}$ is a regularization constant,
$W_{r} \in \mathbb{R}^{{d_u} \times {d_u}}$ is the trainable parameter to transform the neighborhood node $o$ with relationship $r$.

\subsection{Causal Classifier}\label{cause_enc}
We concatenate the final represent representations of the target utterance $u_t$ and the other utterance $u_o$,
then employ MLP with a single hidden layer and a sigmoid function to yield logits $\hat{p}_{o, t}$ and use cross-entropy loss function to do binary classification:
\begin{align}
\hat{p}_{o, t} &= \sigma\left(\textrm{MLP}\left(h_{o}^{'} \oplus h_{t}^{'}\right)\right).
\end{align}

\section{experiment}

\subsection{Experiment Setting}

\begin{table}[t]
\centering
\begin{tabular}{c|r|r|r|r|r}
\hline
Test set & \multicolumn{1}{c|}{Conv.} & \multicolumn{1}{c|}{Utt.} & \multicolumn{1}{c|}{Avg.} & \multicolumn{1}{c|}{Pos. Pairs} & \multicolumn{1}{c}{Neg. Pairs} \\ \hline
DD       & 225                        & 2,405                      & 10                        & 1,894                            & 26,814                           \\ \hline
IE       & 16                         & 665                       & 41                        & 1,080                            & 11,305                           \\ \hline
\end{tabular}
\caption{The statistics of RECCON test set, where "DD" and "IE" stands for the
RECCON-DD and RECCON-IE test sets, respectively.
Avg. represents the number of utterances per conversation on average.}
\label{tab:Statistics}
\end{table}

\noindent\textbf{Dataset.}
We use a recently proposed benchmark dataset for ${\rm C}_2{\rm E}_2$, namely RECCON \cite{poria2021recognizing},
where each utterance in the conversations is attached with a human-annotated emotion label.
It is sampled from two popular datasets of emotion recognition task in conversation~\cite{busso2008iemocap,li2017dailydialog}.
To comprehensively evaluate the model, there are two test sets in RECCON: RECCON-DD
and RECCON-IE.
It is worth noting that the data source of the RECCON-DD is the same as those in the training set, while RECCON-IE is not.
In other words, the RECCON-IE stands for a demanding cross-domain evaluation.  
Specifically, as shown in Table \ref{tab:Statistics},
the average length of each conversation in IE is longer than in DD.
Besides, the emotional shifts in IE conversations are more frequent than in DD, which requires a more complex emotional understanding capacity of the model \cite{poria2021recognizing}.
Following Poria et al. \cite{poria2021recognizing}, we omit those rare-appeared future causes and adopt the ``Fold-1'' of RECCON as the negative samples. 

\smallskip
\noindent\textbf{Implementation Details.}
For training, we choose the cross entropy loss function and set the learning rate to 3e-5 with a batch size of 4.
The dimension $d_u$ of utterance representations and hidden size of MLP are set to 300,
and the number of attention heads and window size are set to 6 and 3, respectively.
For the hyperparameter $c_{t,r}$, we fix it to 2.

\subsection{Results}

\begin{table*}[ht]
\centering
\scalebox{1}{
\setlength{\tabcolsep}{2.5mm}{
\begin{tabular}{ l | l  l  l | l  l  l }
\hline  
\multicolumn{1}{c|}{\multirow{2}{*}{Model}} & \multicolumn{3}{c|}{DD} & \multicolumn{3}{c}{IE} \\
\cline{2-7} 
\multicolumn{1}{c|}{}                       & \multicolumn{1}{l}{Neg. F1} & \multicolumn{1}{l}{Pos. F1} & \multicolumn{1}{l|}{Macro F1} & \multicolumn{1}{l}{Neg. F1} & \multicolumn{1}{l}{Pos. F1} & \multicolumn{1}{l}{Macro F1} \\
\hline
Base\cite{poria2021recognizing}        & 88.74 & 64.28 & 76.51 & 95.67 & 28.02 & 61.85                       \\
ECPE-MLL\cite{ding2020end}    & 94.68 & 48.48 & 71.59 & 93.55 & 20.23 & 57.65                        \\
ECPE-2D\cite{ding2020ecpe}     & 94.96 & 55.50 & 75.23 & \textbf{97.39} & 28.67 & 63.03                         \\
RankCP\cite{wei2020effective}      & \textbf{97.30} & 33.00 & 65.15 & 92.24 & 15.12 & 54.75                          \\
${\rm KEC}^{\clubsuit}$ \cite{li2022neutral}         & 95.74\tiny{($\pm$0.05)} & 66.76\tiny{($\pm$0.33)} & 81.25\tiny{($\pm$0.17)} & 86.08\tiny{($\pm$0.46)} & 19.72\tiny{($\pm$1.71)} & 52.9\tiny{($\pm$0.8)}          \\
\hline
PAGE & 95.80\tiny{($\pm$0.06)} & \textbf{68.80}\tiny{($\pm$0.11)} & \textbf{82.30}\tiny{($\pm$0.05)} & 96.41\tiny{($\pm$0.25)} & \textbf{45.96}\tiny{($\pm$0.82)} & \textbf{71.19}\tiny{($\pm$0.52)}    \\

~~~-w/o PaG  & 93.36\tiny{($\pm$0.46)}  & 52.94\tiny{($\pm$0.97)} & 73.15\tiny{($\pm$0.31)}  & 84.53\tiny{($\pm$2.0)}  & 21.62\tiny{($\pm$0.32)}  & 53.07\tiny{($\pm$0.89)} \\
\hline

\end{tabular}}}
\caption{The results on RECCON. We use Macro F1 as an overall metric, while the Pos. F1 and Neg. F1 represent F1 score on positive and negative pairs, respectively. $\clubsuit$: since KEC \cite{li2022neutral} only report the results on DD, we run the KEC algorithm on the IE test set and report the results. }
\label{tab:result}
\end{table*}

We compare our approach with solid baselines.
As shown in Table \ref{tab:result}, the scores of the first four baselines are reported by Poria et al. \cite{poria2021recognizing}, which are the best-run results among several repeated experiments.
For better comparison, we follow Li et al. \cite{li2022neutral} (i.e., KEC),
reporting the average F1 score and the corresponding variance over five random runs.

Overall, we achieve state-of-the-art performance on both test sets under two main metrics (i.e., Pos. F1 and Macro F1), except the Neg. F1 due to the imbalance in the number of the Pos. and Neg. samples, as shown in Table \ref{tab:Statistics}.
Compared to their methods,
our relative position encoding strategy can fully consider inter-speaker dependency,
which can effectively enhance the understanding of utterances.
Since the conversations in the IE set are longer than those in the DD set, as shown in Table \ref{tab:Statistics}, the emotion cause shifts more in the IE set.
Therefore, detecting the causes in the IE set is more challenging than in the DD set.  
Owing to our PaG structure, which has advantages in aggregating long-term contextual information,
our model still maintains excellent results on IE, with a notable performance gap compared with other baselines.
Since the conversations in real-world applications are mostly verbose \cite{jiao2020real}, this promising performance demonstrates the robustness and practicality of our approach. 

To further prove the effectiveness of our position encoding strategy, we conduct an additional ablation study.
As shown in the bottom row of Table \ref{tab:result}, after removing the PaG, the performance drops significantly, especially in those lengthy conversations of the IE test set.
This result indicates that position is beneficial to emotion cause inference in the conversations,
which is particularly helpful to those long conversations, demonstrating the outstanding generalization capacity and potential practicality of our model.

\subsection{Effect of Window Size}

\begin{figure}[ht]
		\centering
		\includegraphics[width=0.9\linewidth]{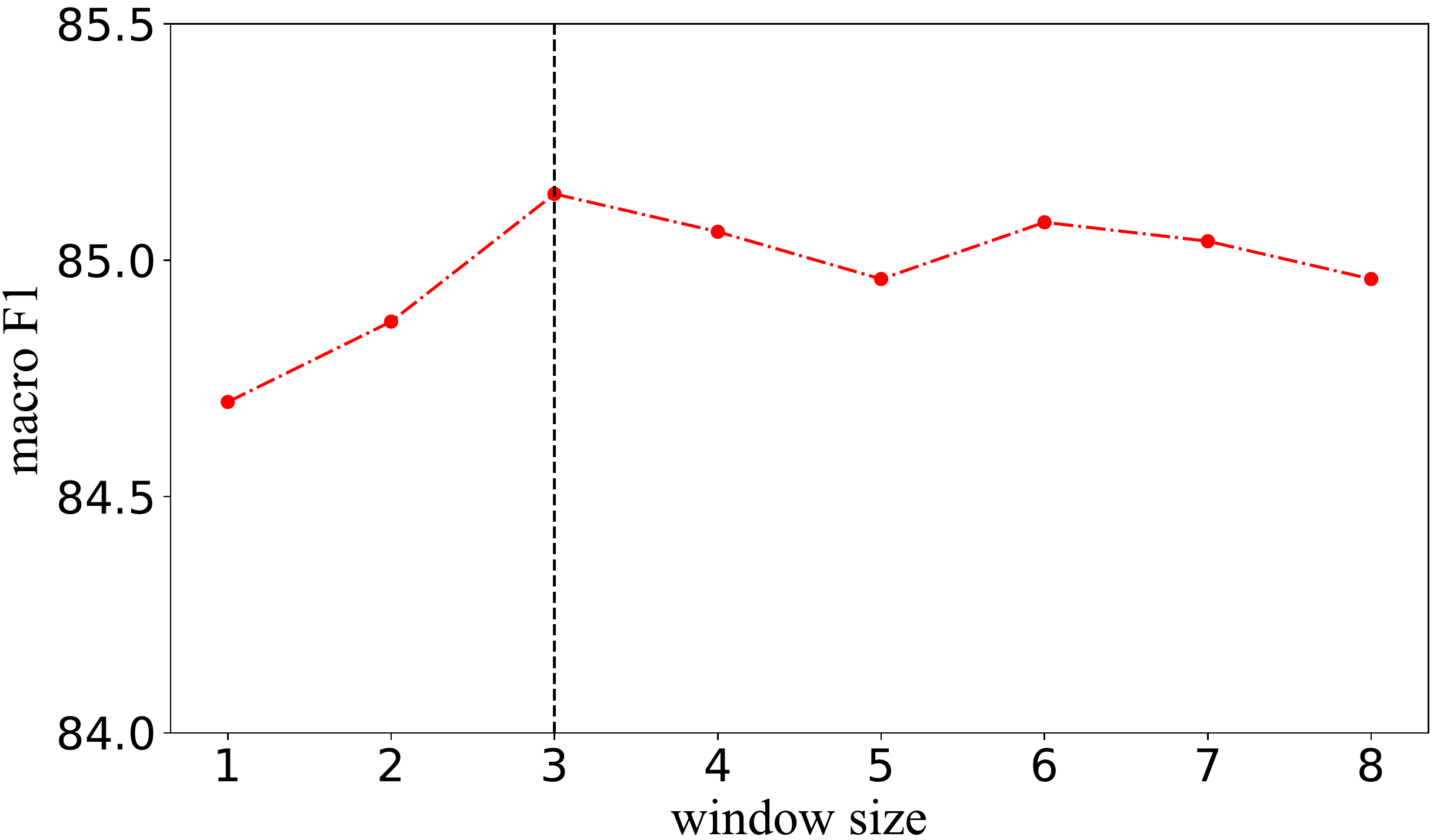}
	\caption{Performance on the DD validation set with varying window size.}
	\label{fig:window}
\end{figure}
The window size may have an underlying effect on the model performance.
Hence, we experiment to investigate the performance fluctuation from various window sizes.
Note that the validation set of RECCON comes from the same source as the DD test set, while there is no corresponding validation set for IE.
Figure \ref{fig:window} presents the results.
Theoretically, a larger window size can lead to more position categories in the graph, but it increases the complexity of the graph network.
While a tiny window represents that only the adjacency of target utterances has an informative position signal.
As shown in Figure \ref{fig:window}, when we increase the window size from 1 to 3, the performance is improved because the diverse position categories benefit the context understanding.
However, the large window size (\(>3\)) has adverse effects due to the inference noise from the long-distance utterances (i.e., causal-irrelevant context).
Thus we choose window size = 3 in the PAGE.

\section{Conclusion}
In this paper, we propose a position-aware graph for ${\rm C}_2{\rm E}_2$ task,
which is beneficial for emotion cause entailment.
Specifically, our framework takes advantage of an effective position encoding strategy incorporating inter-speaker dependency, thus enhancing the capacity for complex emotion cause reasoning. 
Our PAGE model achieves SOTA performance on two challenging test sets,
with 1.1\% and 8.1\% improvement at Macro F1 compared to previous models.

\clearpage
\small
\bibliographystyle{IEEEbib}
\bibliography{strings,refs}

\end{document}